\documentclass[11pt,letterpaper]{article}
\usepackage{naaclhlt2015}
\usepackage{times}
\usepackage{latexsym}

\usepackage[]{graphicx}
\usepackage[english]{babel}
\usepackage[latin1]{inputenc}
\usepackage{lingmacros}
\usepackage{url}
\usepackage{subfigure}
\usepackage{color}
\usepackage{ucs}
\usepackage{enumitem}

\title{GPLSIUA: Combining Temporal Information and Topic Modeling for Cross-Document Event Ordering}

\author{Borja Navarro-Colorado and Estela Saquete\\
        Natural Language Processing Group\\
        University of Alicante\\
        Alicante, Spain\\
        {\tt borja@dlsi.ua.es, stela@dlsi.ua.es}
    }

\date{}

\begin{document}
\maketitle
\begin{abstract}
Building unified timelines from a collection of written news articles requires cross-document event coreference resolution and temporal relation extraction. In this paper we present an approach event coreference resolution according to: a) similar
temporal information, and b) similar semantic arguments. Temporal
information is detected using an automatic temporal information
system (TIPSem), while semantic information is represented by means
of LDA Topic Modeling. The evaluation of our approach shows that it
obtains the highest Micro-average F-score results in the SemEval-2015 Task 4: ``TimeLine: Cross-Document Event Ordering'' (25.36\% for TrackB, 23.15\% for SubtrackB), with an improvement of up to 6\% in
comparison to the other systems. However, our experiment also showed some drawbacks in the Topic Modeling approach that degrades performance of the system.

\end{abstract}

\section{Introduction}
Since access to knowledge is crucial in any domain, connecting and
time-ordering the information extracted from different documents is
a very important task. The goal of this paper is therefore to build
ordered timelines for a set of events related to a target entity. In
doing so, our approach is dealing with two problems: a)
cross-document event coreference resolution and b) cross-document
temporal relation extraction.

In order to arrange event mentions in a timeline it is necessary to
know which event mentions co-refer to the same event or fact and
occur at the same moment. Our approach attempts to formalize the
idea that two or more event mentions co-refer if they have not only
temporal compatibility (the events occur at the same time) but also
semantic compatibility (the event mentions refers to the same facts,
location,  entities, etc.).

Of a set of event mentions in one or more texts, our proposal groups together the event mentions that (i) have the same or a similar temporal reference, (ii) have the same or a similar event head word, and (iii) whose main arguments refer to the same or similar topics. In order to evaluate the system, we have participated in the SemEval-2015 Task 4 ``TimeLine: Cross-Document Event Ordering''.

In the following sections we will present the theoretical background to our approach (section \ref{background}) and the main technical
aspects (sections \ref{approach} and \ref{architecture}). Then we will present the results obtained (section \ref{results}) and some conclusions.

\section{Background}\label{background}

Two or more event mentions co-refer when they refer to the same real
fact or event. Two events can denote the same fact whereas the
linguistic mentions have a different syntax structure, different
words, or even a different meaning.
Whatever the case may be, both event mentions must be semantically
related.

An event mention is formed of an event head (usually a verb or a
deverbal noun) that is related to a semantic structure
(linguistically represented as an argument structure with an agent,
patient, theme, instrument, etc., that is, the semantic roles) in
which there are some event participants (entities)  and which is
located in place and time \cite{Levin2005,Hovav2010}. The
meaning of an event mention is therefore not only the meaning of the
event head, but also the compositional meaning of all the components
and their relations: head, participants, time, place, etc.

In order to detect this semantic relation between event mentions,
previous papers have isolated the main components of the event
structure. For instance, Cybulska and Vossen
\shortcite{conf/ranlp/CybulskaV13} apply an event model based on
four components: location, time, participant and action. Moreover, with regard to
temporal information, only explicit temporal expressions that
appears in the text are considered, but no temporal information is
inferred by navigating temporal links. Bejan and Harabagiu \shortcite{Bejan2014} use a rich set of
linguistic features to model the event structure, including lexical
features such as head word and lemmas, class features such as PoS or
event class, semantic features such as WordNet sense or semantic
roles frames, etc. They use an unsupervised approach based on a
non-parametrical Bayesian model.

\section{Our Approach}\label{approach}

In our approach we represent each event mention as a head word (the
event tag in the TimeML \cite{TimeMLguide} annotation scheme)
related to a temporal expression (implicit or explicit), a set of
entities (0 or more), and a set of topics  that represents what the
event mention is referring to. This paper is focused on temporal
information processing and topic-based semantic representation.

\subsection{Temporal Information Processing}

The TimeML \cite{TimeMLguide} annotation scheme has now been adopted
as a standard by a large number of researchers in the field of
temporal information annotation. It represents not only events and temporal expressions, but also links \cite{Pustejovsky2003TimeML}

A manual annotation of event mentions and the DCT of texts have been
considered as an input of the system, and an automatic system has
been used to perform the annotation with temporal expressions and
temporal links in order to be able to establish a complete timeline
of the input texts. If a plain text is considered, systems such
TIPSem (\underline{T}emporal \underline{I}nformation
\underline{P}roces\-sing using Semantics)
\cite{Llorens2013-IPM-ENTITIES,Llorens2012-IJIS-LINKS}\footnote{\url{http://gplsi.dlsi.ua.es/demos/TIMEE/}}
are able to automatically annotate all the temporal expressions
(TIMEX3), events (EVENT) and links between them.

Once the temporal links have been established, all the specific
temporal information for each event is inferred by means of temporal
links navigation. This information allows us to determine temporal
compatibility between all the events considered.

\subsection{Topic-based Semantic Representation}

The meaning of each event structure has been represented by using Topic Modeling \cite{Blei2012} on a reference
corpus. Topic modeling is a family of algorithms that automatically discover
topics from a collection of documents. More specifically, we apply
the Latent Dirichlet Allocation (LDA) \cite{Blei2003}, which follows
a bottom up approach. Each word is assigned to a topic according to
the co-ocurrence words in the context (document) and the topics
assigned to this word in other documents. In formal terms, a topic
is a distribution on a fixed vocabulary.
We have applied the LDA to the WikiNews
corpus.\footnote{\url{https://dumps.wikimedia.org/enwikinews/}} Each
topic in this corpus is represented using the twenty most prominent
words.
\section{Architecture of the System}\label{architecture}

Our approach to build timelines from written news in English implies
event coreference resolution by applying three cluster processes in
sequential  order: a temporal cluster, a lemma cluster, and a topic
cluster. It combines various resources:

\begin{itemize}[noitemsep,nolistsep]
\item Named entity recognition, using OpeNER web services.\footnote{\url{http://www.opener-project.eu/webservices/}}
\item TimeML automatic annotation of texts using TipSEM system \cite{Llorens2010TempEval-2}.
\item The NLTK\footnote{\url{http://www.nltk.org/}} verb lemmatizer based on WordNet \cite{Fellbaum1998}.
\item The SENNA \cite{SENNA} Semantic Roles Labeling.
\item The LDA Topic Modeling algorithm, using MALLET \cite{McCallumMALLET}.
\end{itemize}

\subsection{Target Entity Filtering}

If the target entity filtering is to be performed then it is first
necessary to resolve the named entity recognition and coreference
resolution. This is done by integrating the external OpeNER web
services into our proposal. More specifically, the components
applied in our proposal are the NER component,\footnote{\url{http://opener.olery.com/ner}} which identifies the
names of people, cities, and museums, and classifies them in a
semantic class (PERSON, LOCATION, etc.) and the coreference
resolution component,\footnote{\url{http://opener.olery.com/coreference}} whose
objective is to identify all those words that refers to the same
object or entity.

Only those events that are part of sentences containing the target
entity or a coreference entity of the target will be selected for
the final timeline.

\subsection{Temporal Clustering Approach}

A plain text was considered and we use the TIPSem system to
automatically annotate all the temporal expressions (TIMEX3), events
(EVENT) and links between them. The TLINKS annotated in the text are
used in order to extract the time context of each event and make it
possible to infer both time at which each event occurs and the
temporal ordering between the events in the text. Moreover, if we
are able to determine the time of the event, we will be able to
determine temporal compatibility between events, even when they are
contained in different documents, thus signifying that
cross-document event coreference resolution is also possible.

In this first step, all the events from the different documents that
occurring on the same date will therefore be part of the same
cluster. The clusters are positioned in ascending ordered based on
the date assigned.

\subsection{Semantic Clustering Based on Lemmas}
Once all the events that share temporal information and the target
entity have been grouped together, we apply a simple clustering
based on head word lemmas. This lemma-based clustering groups
together all event mentions with the same head word lemma, the same
temporal information and the same target entity. We therefore assume
that all these event mentions corefer to the same event. This is our Run 1 at the competition.

\subsection{Semantic Clustering Based on Topics}

The problem of the lemma-based cluster is that it does not take into
account the argument structure of the event. This last clustering
therefore attempts to solve this problem by extracting the semantic
roles from each event and representing their meaning by using topics
on a reference corpus. This approach has three steps:

\begin{enumerate}[noitemsep,nolistsep]
\item Using SENNA \cite{SENNA} as Semantic Roles Labeling, we have detected roles A0 and A1.\footnote{In order to represent Semantic Roles, SENNA uses the tag set proposed by Proposition Bank Project (\url{http://verbs.colorado.edu/~mpalmer/projects/ace.html}) A0 and A1 represent the main roles related to each verb.} which are related to the event mention head word. For each role we extract only the nouns.

\item We have extracted 500 topics from WikiNews using Topic Modeling with MALLET. All these topics are used as a knowledge base. We will use only the most representative words for each topic (the twenty words with the greatest weight) and the weights that they have in each topic.

\item Finally, we have created an event-topic matrix. Each event (raws) is represented by a vector. The values of the vector are the  addition of weights of each argument noun in each topic (columns).
\end{enumerate}

For example, if the nouns in arguments A0 and A1 are ``users,
problems, phones'', we represent their meanings according to the
topics $t_n$ assigned to them by applying LDA to WikiNews ($user = {t_0,t_3,t_5}$, $problems = {t_0,t_2}$, $phones = t_5, t_6$, etc). Then, the event $e$ of this sentence is represented by a $n$-dimensional vector in which $n$ is the amount of topics (500) and whoses values are the addition of weight of each noun in each topic $T_n$.

In order to group together similar event mentions, we have applied a
k-means clustering algorithm to these event vectors.\footnote{Note that it has been applied only to the events previously clustered following the lema-based approach (Run 1).} The distance metric used has been Euclidean Distance. The number of cluster has been adjusted to two.\footnote{We have used PyCluster tool: \url{https://pypi.python.org/pypi/Pycluster}} Therefore, each cluster with the
same head word lemma, the same temporal information and the same
target entity is then re-clustered according to the similarity of
the main topics of its arguments. This cluster corresponds to our Run 2 at the competition.

\section{Evaluation Results}\label{results}
SemEval-2015 Task 4 consists on building timelines from written news
in English in which a target entity is involved. The input data
provided by the organizers is therefore a set of documents and a set
of target entities related to those documents. Two different tracks
are proposed in the task, along with their subtracks:

\begin{itemize}[noitemsep,nolistsep]
\item Track A: This consists of using raw texts as input and obtaining full timelines. Subtrack A has the same input data, but the output will be the timeLines of only ordered events (no assignment of time anchors).
\item Track B: This consists of using texts with manual annotation of events mentions as input data. Subtrack B has the same input data but the output will be timeLines of only ordered events.
\end{itemize}

In the Semeval-2015 Task 4 competition we have participated in Track
B and Subtrack B. The results for the Micro-average F-score measure obtained by our approach in
the competition are shown in Table \ref{resultsGPLSIUA}.

\begin{table} [htbp]
\begin{footnotesize}
\begin{center}
\begin{tabular} {lcccc}
\hline 
\textbf{TRACK} & \textbf{Corpus1} & \textbf{Corpus2} & \textbf{Corpus3} & \textbf{Total}\\
  \hline\rule{-2pt}{7pt}
 TrackB-R1 & 22.35
& 19.28 & 33.59 & \textbf{25.36}\\
 TrackB-R2 & 20.47
& 16.17 & 29.90 & \textbf{22.66}\\
 SubTrackB-R1 & 18.35
& 20.48 & 32.08 & \textbf{23.15}\\
 SubTrackB-R2 & 15.93
& 14.44 & 27.48 & \textbf{19.18}\\
  \hline
\end{tabular}
\caption{Results for GPLSIUA Approach.}\label{resultsGPLSIUA}
\end{center}
\end{footnotesize}
\end{table}

Although the Micro-FScore results are not very high, the results obtained by our approach are the highest in all of the
corpus evaluated by the organizers. Our approach obtained an improvement of 7\% compared with the other participant in Track B and a 6.48\% in Subtrack B.

\section{Conclusions}

The results show that our approach is suitable for the task in hand. On the one hand, temporal
information is automatically extracted with a temporal information
processing system which makes it possible to infer and determine the
time at which each event has occurred. On the other hand, the
semantic similarity based on the verb is sufficient to group
together coreferent events.

The basic method (Run 1), consisting of searching for similar verb
lemma, eventually proved to be the best. We have therefore carried
out an in-depth analysis of the results obtained for Run 2 and have
observed three main drawbacks in the Topic Modeling approach:

\begin{itemize}[noitemsep,nolistsep]
\item The K-means algorithm forces us to fix the number of clusters beforehand, and this has been fixed at 2. However, there is often only one correct cluster. Another approach without a fixed number of topics will improve the approach. Bejan and Harabagiu \shortcite{Bejan2014}, for example, suggest inferring this value from data.
\item The representativity of each event mention depends directly on the amount of topics extracted from the reference corpus. Many topics will produce excessive granularity, and few topics will be unrepresentative. We have set the number of topics at 500, but it is necessary to study whether another amount of topics will improve the results.

\item This approach depends excessively on the representativity of the reference corpus. We believe using larger corpora should improve the results.
\end{itemize}

As Future work, we plan to use other similarity measures and
clustering algorithms in an attempt to solve the problem of
previously fixed number of clusters. We also plan to evaluate using different Topic Modeling configurations.

\section*{Acknowledgments}
\begin{footnotesize}
We would like to thank the anonymous reviewers for their helpful suggestions and comments.
Paper partially supported by the following projects: ATTOS (TIN2012-38536-C03-03), LEGOLANG-UAGE (TIN2012-31224), SAM
(FP7-611312), FIRST (FP7-287607) DIIM2.0 (PROMETEOII/2014/001)
\end{footnotesize}
\bibliographystyle{naaclhlt2015}
\bibliography{biblio}

\end{document}